# A NOVEL INTRINSIC IMAGE DECOMPOSITION METHOD TO RECOVER ALBEDO FOR AERIAL IMAGES IN PHOTOGRAMMETRY PROCESSING

Shuang Song [1,2], Rongjun Qin [1,2,3,4,*]

[1] Geospatial Data Analytics Laboratory, The Ohio State University, Columbus, USA
[2] Department of Civil, Environmental and Geodetic Engineering, The Ohio State University, Columbus, USA
[3] Department of Electrical and Computer Engineering, The Ohio State University, Columbus, USA
[4] Translational Data Analytics Institute, The Ohio State University, Columbus, USA
Email: <song.1634, qin.324>@osu.edu

**Commission II, WG II/1**

**KEY WORDS:** Photogrammetry, Aerial Imagery, Albedo, Illumination, Intrinsic Image Decomposition, Shadow Matting

**ABSTRACT:**

Recovering surface albedos from photogrammetric images for realistic rendering and synthetic environments can greatly facilitate its downstream applications in VR/AR/MR and digital twins. The textured 3D models from standard photogrammetric pipelines are suboptimal to these applications because these textures are directly derived from images, which intrinsically embedded the spatially and temporally variant environmental lighting information, such as the sun illumination, direction, causing different looks of the surface, making such models less realistic when used in 3D rendering under synthetic lightings. On the other hand, since albedo images are less variable by environmental lighting, it can, in turn, benefit basic photogrammetric processing. In this paper, we attack the problem of albedo recovery for aerial images for the photogrammetric process and demonstrate the benefit of albedo recovery for photogrammetry data processing through enhanced feature matching and dense matching. To this end, we proposed an image formation model with respect to outdoor aerial imagery under natural illumination conditions; we then, derived the inverse model to estimate the albedo by utilizing the typical photogrammetric products as an initial approximation of the geometry. The estimated albedo images are tested in intrinsic image decomposition, relighting, feature matching, and dense matching/point cloud generation results. Both synthetic and real-world experiments have demonstrated that our method outperforms existing methods and can enhance photogrammetric processing.

## 1. INTRODUCTION

### 1.1 Introduction

Today, aerial photogrammetric 3D reconstruction from images has been developed into mature toolchains such as for lightweight unmanned aerial vehicle systems (UAVs) and commercial / open-sourced data processing software, which can generate high resolution, high accuracy, and photorealistic textured models for various applications (Alidoost and Arefi, 2017). However, the demands on data quality for novel applications are growing even faster, *e.g.*, virtual reality, augmented reality, mixed reality, and the metaverse. It was noted that the quality of photogrammetric geometry has been recognized, but the textures do not meet expectations (Innmann et al., 2020; Lachambre, 2018). The reason is that classic aerial photogrammetry tends to supply texture materials directly from the images as "albedos" as required by graphics rendering pipelines, which, however, is drastically different: for example, an albedo texture material is free of shadows so the graphics rendering pipeline can cast shadows in a simulated environment, while the image textures may already possess unwanted cast shadows by the actual environmental lighting. On the other hand, we are also curious if albedo images can be recovered, and how it may enhance or impact the photogrammetric processing. Therefore, in this paper, we aim to study the problem of albedo recovery and its application to photogrammetric data processing.

The concept of deriving albedo from images was widely used in the remote sensing community with applications to satellite imagery (Haest et al., 2009; Honkavaara et al., 2013, 2012; O'Hara and Barnes, 2012). Various products (e.g., MODIS BRDF/Albedo products) were produced and used in environmental and remote sensing studies (Pisek et al., 2016, 2015; Z. Wang et al., 2018). Nonetheless, those models only assume a 2D/2.5D world and are not suitable for close-range photogrammetry.

Intrinsic image decomposition, which originated in compute vision, studies the problem of recovering albedo and illumination from a single image, it has been investigated since the Retinex theory (Land and McCann, 1971) was proposed in 1971. Most methods assume a slow varying shading which can be modeled with smoothing PDE (partial differential equations). The latest works involve deep learning models to learn albedo recovery from the huge volume of training data (Li and Snavely, 2018; Sheng et al., 2020; Yu and Smith, 2021, 2019). However, those methods often operate in black-box mode and lack generalization and the ability to comprehend complex physical lighting models in outdoor scenarios (Sheng et al., 2020).

The cast shadow is a common effect in aerial imagery under a clear sky condition which yields a strong discontinuity in shading. The goal of shadow detection and removal is to localize masks of cast shadow in an image and compensate pixels to produce shadow-free images (Mostafa, 2017; Sasi and Govindan, 2015). Nonetheless, only the high contrast cast shadows are of concern

---

but the smooth shading is overseen. The deep learning variants of those methods trend toward compensating all shading effects, but it still unstable and less comprehensive (Qu et al., 2017).

In this paper, combining principles of intrinsic image and shadow detection, we propose a method that utilizes *known geometry*, *RAW photos* (digital negative), and their meta-information (*capture date & time*, *GPS location*) to model the image formation process, which addresses sharp and high-contrast shading of *cast shadow* in datasets with a photometric model involving sun and sky irradiance and surface normals (Section 2). Then we develop a method to invert the process to estimate the albedos of the scene (Section 3). The proposed method is evaluated *w.r.t.* various photogrammetric applications, including relighting rendering, feature matching, and dense matching.

### 1.2 Related Works

**Intrinsic image** studies the decomposition of a single image to intrinsic layers (Barrow H.G. and Tenenbaum, 1978) including but not only diffuse albedo (or reflectance) and shading (or illumination). The decomposition is an ill-posed problem that heavily relies on priors to regularize the solution. A few works ask users to provide guidance strokes to regularize the problem (Bousseau et al., 2009; Shen et al., 2011). Most methods assume a sparse or piecewise constant albedo (Gehler et al., 2011) and smooth monochromatic illumination. Automatic methods are built on the consensus that large derivatives of images are attributed to the changes in albedo while small derivatives to shading. Recent works are challenging the assumption by detecting a shadow layer (Sheng et al., 2020) or estimating complex shading with neural networks (Innamorati et al., 2017; Janner et al., 2017; Yu and Smith, 2021, 2019). Physical shading is proposed to further recover surface normal and lighting (Barron and Malik, 2015, 2012a, 2012b, 2011), but they are limited to convex, non-occluded, single objects. By integrating inverse rendering techniques, (Laffont et al., 2013) works on multi-view images of the outdoor scene by simulating light transmission with the ray tracing engine. The method is compatible with complex geometry but requires a few manual setups to collect environment irradiance with a light probe. (Duchêne, 2015; Duchêne et al., 2015) utilized sky and sun portion of ground collected images to estimate environment irradiance, however, it's difficult to obtain looking up images for aerial imaging. Our method does not rely on direct observation of the sun and sky, alternatively, we estimate the essential illumination parameters from observed images.

**Shadow detection and removal** studies the extraction and removal of cast shadows from a single image. Early works including automatic methods (Finlayson et al., 2004) only work on images of uncluttered scenes with isolated shadows. Recent studies often train neural networks to perform end-to-end shadow detection and removal and have achieved great success on datasets (Cun et al., 2020; Qu et al., 2017; J. Wang et al., 2018), but generalizability is of great concern in practice. Constraint-based methods require user-assistant or classifier to indicate the shadow region (Guo et al., 2013; Wu et al., 2007). Photogrammetry and remote sensing community shows great interest in shadow detection removal since for most applications (*e.g.*, dense matching, orthorectification, semantic segmentation) they are considered a nuisance (Donne and Geiger, 2019; Luo et al., 2018; Silva et al., 2018; Wang et al., 2017; Wu and Bauer, 2013). Shadow removal with known DSM (Digital Surface Model) and sun position was proved to be efficient to supply as a constraint (Wang et al., 2017). Pairs of lit and shadow points sharing the same albedo have been adopted by works to estimate the shadow boundaries (Duchêne, 2015; Duchêne et al., 2015). Similar to the works, pairs of lit and shadow points are heavily used in this paper, not only for shadow detection but also for environmental illumination estimation.

## 2. IMAGE FORMATION MODEL

The image formation model describes the physical process of interactions among the light source, surface material, geometry, and the observed image, in which our desired albedos are involved.

### 2.1 Image Model

Our model follows a few basic assumptions:
1) Lambertian surface assumption, which means only diffuse albedo is considered;
2) The intensity of the image is proportional to the irradiance of the corresponding object, which is implied in Equation 1. The RAW format (digital negative) image meets this requirement, which is supported by most cameras and some mobiles.

$$\boldsymbol{I} = \boldsymbol{R} \otimes \boldsymbol{S}, \quad (1)$$

where $\boldsymbol{I} \in \mathbb{R}^3$ is the radiance of a surface point observed by the camera (*i.e.*, the pixel intensity of the image), $\boldsymbol{R} \in \mathbb{R}^3$ is albedo (or reflectance) of the object point, $\boldsymbol{S} \in \mathbb{R}^3$ is the shading (or illumination reached to that object point), $\otimes$ is an element-wise product of two vectors. To be noted, in this paper, vectors and matrices are denoted with the bold symbol.

We then decompose $\boldsymbol{S}$ by the outdoor and natural light assumption following previous works (Duchêne et al., 2015; Laffont et al., 2013), which assumes three components: irradiance from the sun (denoted as $\boldsymbol{S_{sun}}$), sky (denoted as $\boldsymbol{S_{sky}}$) and reflected by other objects (denoted as $\boldsymbol{S_{ind}}$). The Equation 2 shows the image formation with albedo and shadings. Figure 1 illustrates the physical process of radiance accumulation.

$$\begin{aligned}\boldsymbol{I} &= \boldsymbol{R} \otimes (\boldsymbol{S_{sun}} + \boldsymbol{S_{sky}} + \boldsymbol{S_{ind}}) \\ &= \boldsymbol{R} \otimes (\boldsymbol{L_{sun}}\alpha_{sun}cos(\omega_{sun}) + \boldsymbol{S_{sky}} + \boldsymbol{S_{ind}}),\end{aligned} \quad (2)$$

where $\alpha_{sun} \in [0,1]$ is the sun visibility factor from the point, $\alpha_{sun} = 1$ where the point is fully lit by the sun, or $\alpha_{sun} = 0$ when located in the umbra, and other intermediate values if located in the penumbra. $\boldsymbol{L_{sun}} \in \mathbb{R}^3$ is the irradiance of the sun, $\omega_{sun}$ is the angle between the surface normal $\boldsymbol{n} \in \mathbb{R}^3$ ( $\|\boldsymbol{n}\|_2 = 1$ ) and the sun position $\boldsymbol{\theta_{sun}} \in \mathbb{R}^3$ ( $\|\boldsymbol{\theta_{sun}}\|_2 = 1$ ), $\boldsymbol{S_{sky}} \in \mathbb{R}^3$ is the irradiance of the sky hemisphere and $\boldsymbol{S_{ind}} \in \mathbb{R}^3$ is the irradiance of indirect illumination.

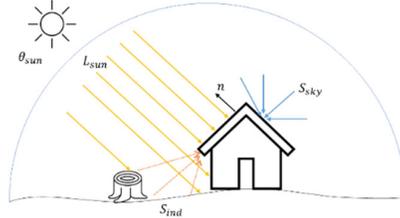

**Figure 1.** The outdoor illumination model (Yellow arrows are sun illumination; blue arrows are sky illumination; red arrows are reflected indirect illumination).

As we can notice that $\boldsymbol{S_{sky}}$ and $\boldsymbol{S_{ind}}$ for each point are integral terms that are defined over a hemisphere, which requires predefined/known environmental radiance. Existing solutions usually record environmental radiance with a light probe while capturing data. (Duchêne, 2015; Duchêne et al., 2015) avoid light probe capturing by reusing sky pixels already captured by images in the dataset to approximate environmental radiance. In this paper, we do not rely on known environmental radiance or direct observation from external sources, because our model enables the estimation of the essential lighting parameters from the dataset (*i.e.*, aerial images) (Section 3.2).

## 2.2 Approximation of Indirect Illumination

Indirect illumination is an effect of the reflection of diffuse surfaces of the scene. It is costly to model and challenging to integrate into inverse rendering due to its non-linearity. Fortunately, we demonstrate that indirect illumination has minor effects in the outdoor scenario. Figure 2 presents two images rendered without (left) and with indirect illumination. By both visual comparison and evaluating PSNR of two images, they indicate that the impact of ignoring indirect illumination is limited. In our model, considering the trade-off between costs and benefits of indirect illumination, we force $S_{ind} = 0$ to make our method concise.

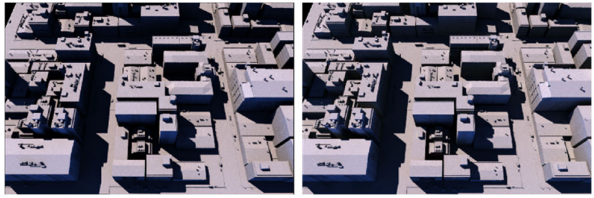

(a) wo/ indirect illumination  (b) w/ indirect illumination
**Figure 2.** Evaluate the impact of indirect illumination, the PSNR for image (a) with (b) as the reference is 76.75dB.

## 2.3 Approximation of Sky Illumination

We proposed the following sky illumination term to approximate the integral term with the assumption that sky radiance is uniformly distributed.

$$S_{sky} = L_{sky}(0.5 + 0.5\,cos(\omega_{zenith}))\ ,\quad (3)$$

where $L_{sky} \in \mathbb{R}^3$ is irradiance of the sky hemisphere, $\omega_{zenith}$ is the angle between the surface normal $n \in \mathbb{R}^3$ and zenith direction $\theta_{zenith} \in \mathbb{R}^3$, in most cases $\theta_{zenith} = (0,0,1)^T$.

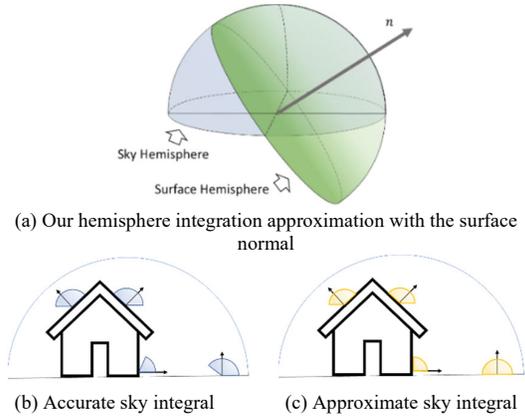

(a) Our hemisphere integration approximation with the surface normal

(b) Accurate sky integral  (c) Approximate sky integral
**Figure 3.** Our sky illumination model.

To simplify this model, we assume no multi-path reflection in the aerial case, since those effects are rather weak in outdoor scenarios whose lighting is dominated by the Sun (Section 2.2).

As Figure 3 (b) and (c) show, our model can approximate most outdoor scenarios, especially, the top and façade of buildings. The fitting error becomes larger while closing to the building, due to the ambient occlusion effect. Finally, our image formation model is written as Equation 4. In our problem, $I, \omega_{sun}, \omega_{zenith}$ are given data, we will estimate the $\alpha_{sun}, L_{sun}, L_{sky}$ in Section 3 proposed method.

$$\begin{aligned}I &= R \otimes \big(L_{sun}\alpha_{sun}cos(\omega_{sun}) + L_{sky}(0.5 + 0.5\,cos(\omega_{zenith}))\big) \\ &= R \otimes \big(L_{sun}\alpha_{sun}(\theta_{sun} \cdot n) + L_{sky}(0.5 + 0.5\,\theta_{zenith} \cdot n)\big) \\ &= R \otimes \big(L_{sun}\alpha_{sun}k_{sun} + L_{sky}k_{sky}\big)\end{aligned} \quad,(4)$$

where $k_{sun} = \theta_{sun} \cdot n$ and $k_{sky} = 0.5 + 0.5\,\theta_{zenith} \cdot n$.

## 3. PHYSICAL-BASED INTRINSIC IMAGE DECOMPOSITION IN OUTDOOR SCENARIO

The workflow of our method is shown in Figure 4. The input of our method is a linearized RAW image with EXIF information (capturing date and time, GPS location, etc.). Firstly, we solve the camera intrinsic and extrinsic parameters with a general structure-from-motion (SfM) pipeline and reconstruct dense 3D points with a general multi-view stereo (MVS) pipeline from the image collection. Then, we project geometric attributes (*i.e.*, $\alpha_{sun}, n$) back to images and guide a conditional random field (CRF) to propagate $\alpha_{sun}$ by images. Next, we estimate parameters of illumination models ($L_{sun}, L_{sky}$) and refine $\alpha_{sun}$ simultaneously based on cast shadows. Finally, we assemble the total shading $S$ and decompose albedo with Equation 1.

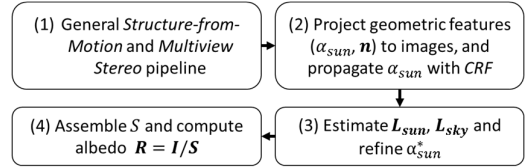

**Figure 4.** The workflow of our intrinsic image decomposition.

### 3.1 Estimate $\alpha_{sun}$ from Geometry and Image

Given the geo-location and image capturing date and time, the local sun position $\theta_{sun}$ can be calculated with an astronomical algorithm (Meeus, 1991). We use a ray-tracing rendering system to initialize $\alpha_{sun}$ by checking the ray visibility toward directional light with $\theta_{sun}$. At meantime, we project the surface normal $n$ of the reconstructed geometry to images.

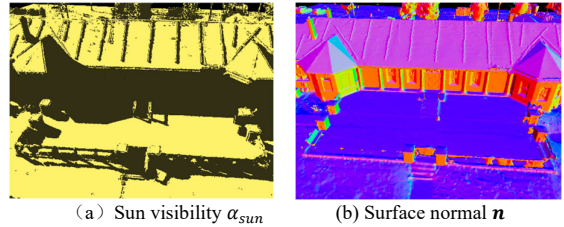

(a) Sun visibility $\alpha_{sun}$  (b) Surface normal $n$
**Figure 5.** Geometric features projection to images.

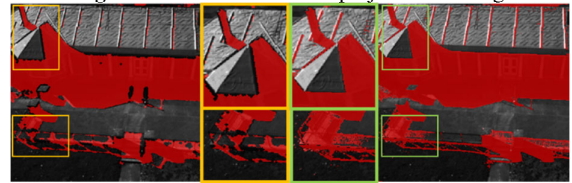

(a) Projected $\alpha_{sun}$  (b) CRF refined $\alpha_{sun}$
**Figure 6.** Propagate sun visibility with image content.

Figure 5 shows sun visibility and surface normal projected to an image plane. Then, we take projected sun visibility as the initial result and refine it with Conditional Random Field (CRF) (Krähenbühl and Koltun, 2011) guided by the image, the comparison before and after the refinement is illustrated in Figure 6.

### 3.2 Estimate $L_{sun}$, $L_{sky}$ with lit-shadow pairs

In this section, we introduce the main idea of estimating illumination factors from observed images based on our image formation model. Estimating $L_{sun}$ and $L_{sky}$ is ill-posed as no absolute gauges of the illumination are known aside from pixel values. Alternatively, we can estimate the relative ratio $L_{sun}/L_{sky}$. As we noticed the natural that pixels across the shadow boundary very likely share the same albedo, the changing of pixel values reflects the difference in illumination. Explicitly, a fully lighted pixel $I_{lit}$ (i.e., $\alpha_{sun} = 1$) and the paired pixel $I_{shadow}$ located in umbra (i.e., $\alpha_{sun} = 0$) are modeled with:

$$\begin{cases} I_{lit} = R \otimes (L_{sun}k_{sun} + L_{sky}k_{sky}) \\ I_{shadow} = R \otimes (L_{sky}k_{sky}) \end{cases}. \quad (5)$$

By assuming albedo $R$, shading coefficients $k_{sun}$, $k_{sky}$ are the same, we can derive the ratio $L_{sun}/L_{sky}$ as Equation 6.

$$\frac{L_{sun}}{L_{sky}} = \frac{I_{lit} - I_{shadow}}{I_{shadow}} \cdot \frac{k_{sky}}{k_{sun}}. \quad (6)$$

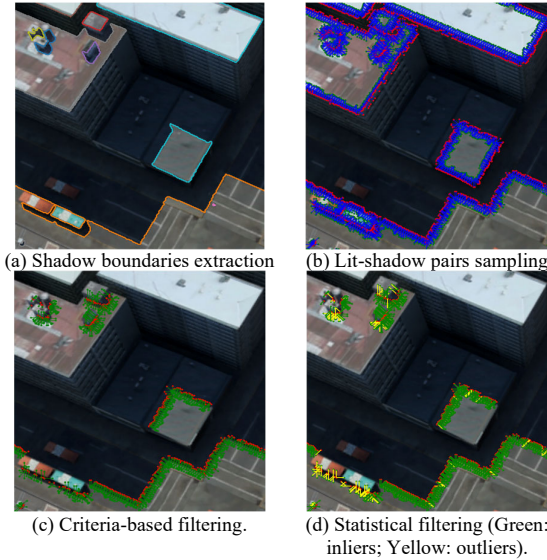

(a) Shadow boundaries extraction  (b) Lit-shadow pairs sampling

(c) Criteria-based filtering.  (d) Statistical filtering (Green: inliers; Yellow: outliers).

**Figure 7.** Find reliable lit-shadow pairs to estimate $L_{sun}/L_{sky}$.

To generate point pairs for estimation, we propose criteria-based filters (Algorithm 1) and a statistical. Firstly, given the sun visibility mask $\alpha_{sun}$, we extract the boundaries of shadow regions (Figure 7(a)) and samples pairs in the lighted and umbra regions (Figure 7(b)). Secondly, since Equation 6 requires the albedos and surface normals of the pair of points are the same, we adapt conditions in Algorithm 1 to eliminate pairs based on geometry and appearance properties Figure 7(c). Finally, we fit a 2-component Gaussian Mixture Model (GMM) for the ratio $L_{sun}/L_{sky}$ computed from remaining pairs as shown in Figure 8(a). If the major component is consists of 95% pairs, and the covariance is significantly smaller than the second component, we accept the mean as the estimation of $L_{sun}/L_{sky}$ (Figure 7(d) and Figure 8(b)). To be noted that for a single image, the estimation could be failed if the shadow boundary is too few, however, the quantity of the $L_{sun}/L_{sky}$ is a constant across the entire fixed illumination dataset. Also, the ratio doesn't effected by exposure parameter. By integrating more pairs across the entire dataset, our method could provide a robust estimation.

---

**Algorithm 1 Criteria-based filters**

1. Remove overexposure and underexposure pixels.
2. Angle of the two surface normals is less than 5°.
3. Depth difference of the two points is less than a threshold.
4. $k_{sky}/k_{sun} \in (0.1, 10.0)$ to ensure numerical stability.

---

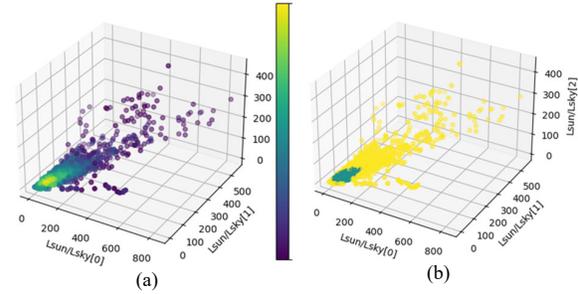

**Figure 8.** (a) Probability distribution estimated by GMM(n=2). (b) Pairs of the major component are inliers (green), and of the minor component are outliers (yellow).

### 3.3 Optimize $\alpha_{sun}$ with Tikhonov regularization

At this stage, all necessary quantities of Equation 4 have been estimated. The albedo component (Figure 9(a)) is resolved with the inverse image model $R = I / S$. Nonetheless, the seamlines between lighted and shadowed regions are sharp because the CRF framework made the binary assumption CRF (denoted as $\alpha_{sun}^0$), either shadow or non-shadow. In this section, we further improved it to a continuous representation to match the nature of penumbra. Analysis of the profile shown in Figure 9(c) indicates the source of spikes in albedo layers. To address the spiky seamline, we build a 1D optimization problem with Tikhonov regularization (Tikhonov, 1963).

$$\alpha_{sun}^*(t) = argmin \frac{1}{2}\|\alpha_{sun}(t) - \alpha_{sun}^0(t)\|_P^2 + \frac{1}{2}\left\|\nabla\left(\frac{1}{R(t)}\right)\right\|^2, (7)$$

with

$$\frac{1}{R(t)} = \frac{L_{sun}k_{sun}(t)}{I(t)}\alpha_{sun}(t) + \frac{L_{sky}k_{sky}(t)}{I(t)}, \quad (8)$$

where $\|\cdot\|_P$ = Mahalanobis distance with weight matrix $P$
$\nabla(\cdot)$ = Gradient operator
$t$ = distance along profile.

In Equation 7, $t$ is the parameter of the 1D signal that represents distance along profile (Figure 9(c) and (d)). For the regularization term, we choose to optimize $1/R$ since compared with directly optimizing regarding $R$, the $1/R$ yields a closed-form solution due to its linearity w.r.t. $\alpha_{sun}(t)$, as shown in Equation 8. Weight matrix $P$ is a diagonal positive definite matrix. We prefer to adjust the $\alpha_{sun}(t)$ that close to shadow boundary but keep the two far ends stationery to ensure a smooth transition to the adjusted region. Figure 9(b) and (d) show the estimated albedo with optimized soft sun visibility $\alpha_{sun}^*$.

Combining $\alpha^*_{sun}$ and $L_{sun}/L_{sky}$, our ultimate solution is shown in Figure 10(a) and (b). Figure 10(c) shows the ground truth shading image computed from rendered image and diffuse albedo image.

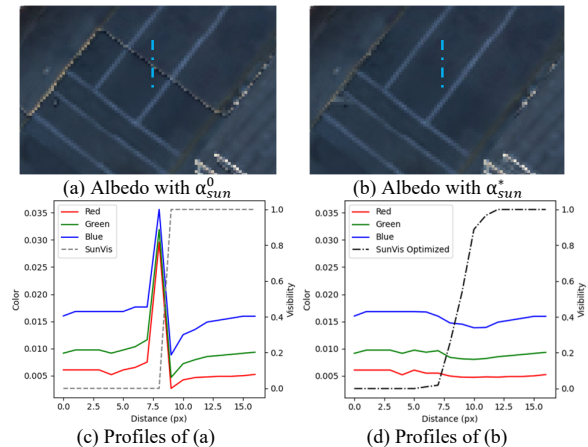

(a) Albedo with $\alpha^0_{sun}$  (b) Albedo with $\alpha^*_{sun}$

(c) Profiles of (a)  (d) Profiles of (b)

**Figure 9.** (a) and (c) are the decomposed albedo with binary $\alpha^0_{sun}$ from CRF that yields spikes in profile; (b) and (d) are the decomposed albedo with refined $\alpha^*_{sun}$ that yields a smooth seamless transition.

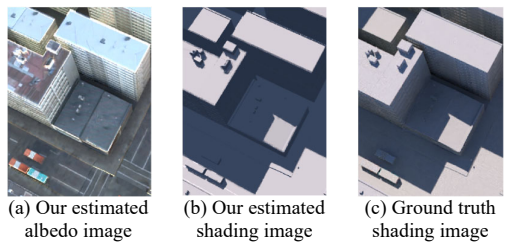

(a) Our estimated albedo image  (b) Our estimated shading image  (c) Ground truth shading image

**Figure 10.** Our albedo and shading estimation.

## 4. EXPERIMENTS

We collect 350 images in both DNG and JPEG formats with DJI Phantom Pro 4 v2.0 for evaluation. Pixels of DNG images are 16bit depth and in linear color space. GPS metadata was extracted from corresponding JPEG files. Then we process images with Agisoft Metashape Professional 1.8.0 (Agisoft, 2022a) yielding image intrinsic & extrinsic parameters, sparse and dense point clouds with normal, as well as a surface model with textures as shown in Figure 11. In the following sections, we evaluate our method from 4 perspectives: intrinsic image decomposition, relighting, feature matching, and dense matching.

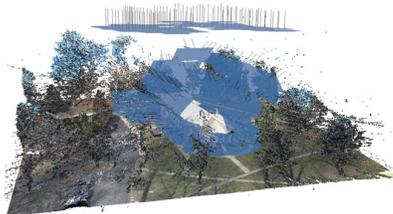

**Figure 11.** The real-world dataset contains 350 images. GSD: 7mm/px, Area: 844 m$^2$, Mission duration: 46min.

### 4.1 Albedo Recovering

We compare our results with those of the state-of-the-art albedo decomposition methods on the real-world dataset, including InverseRenderNet v2 (Yu and Smith, 2021) and SMGAN (Cun et al., 2020). Since ground truth is absent in the real-world dataset, we show the qualitative comparison in Figure 12. Reference methods are both based on deep neural networks and pre-trained models are provided. As indicated by the red circles in Figure 12, InverseRenderNet v2 failed to recover albedo under the shaded region. SMGAN performs better in shadow removal, however, only large shadows are removed but missed shadows of small objects. SMGAN yields inconsistent albedo in the same image as indicated by yellow boxes. Furthermore, we can compare results at image collection level. Due to neural networks only taking a single image as input, the consistency between images is not guaranteed. In our method, the ratio $L_{sun}/L_{sky}$ is a fixed value in an image collection acquired during a short time, and the ratio is irrelevant to exposure settings. We estimate $L_{sun}/L_{sky}$ with lit-shadow pairs from image collection but not a single image, which enables image collection-wide consistency and robustness (*e.g.*, no reliable shadow boundary can be found from some images). In our results, there are a few flaws can be recognized around the edges of the roof, it is because of the inaccurate normal computed from the reconstructed surface model.

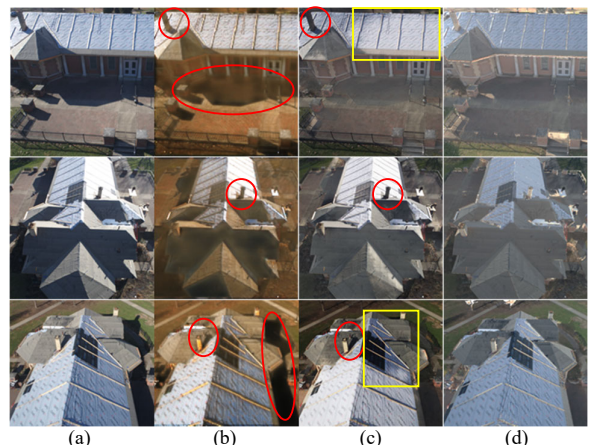

(a)  (b)  (c)  (d)

**Figure 12.** Qualitative comparison of albedo decomposition. (a) Raw image, (b) InverseRenderNet v2, (c) SMGAN, (d) Ours.

### 4.2 Relighting

We evaluate the relighting capability by rendering images from novel views and simulated sky with Blender. We use Agisoft Metashape Professional to create textures from oriented images. The textured model with original images (standard pipeline) is shown in Figure 13(a) and (b). Note that, the left column of Figure 13 is rendered with uniform lighting and the right column is rendered with a different sun position than the collection time. Ideally, under uniform lighting, no shadow or shading effect should be observed. And with a given sun position, the shadow azimuth and shading should be coherent with physical law. The reference model (Figure 13 (c) and (d)) is created by Agisoft Texture De-Lighter (Agisoft, 2022b), which is a close-sourced free software developed by Agisoft LLC. It requires sparse user strokes to indicate light and shadow regions on a textured model to drive the algorithm. Our textures are created by Agisoft Metashape Professional but use our albedo images instead, as shown in Figure 13 (e) and (f).

The original model (Figure 13(a)) preserves all shadows and illumination changes of Lambert's cosine law (Pharr et al., 2016). De-Lighter is managed to remove the shadow of the building and a large portion of shading, but the software assigned the incorrect color to the object indicated by the yellow boxes. And missed shadows also created artifacts in Figure 13(d). Our model shows better relighting since there are no artifacts of shadow at all. From the perspective of color fidelity, our relighted model shows visually better results than the reference model. This is further validated from a closer view as shown in Figure 14.

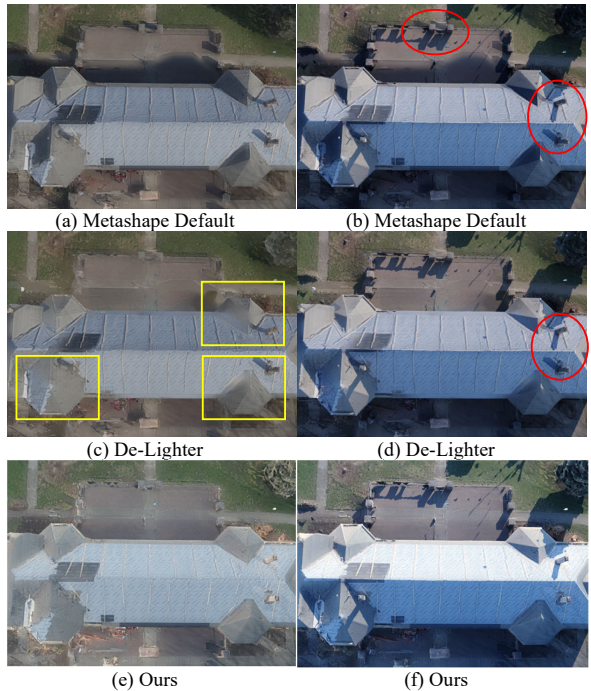

**Figure 13.** Uniform lighting (left column) and sun-sky relighting (right column). From top to bottom: (a) and (b) standard pipeline; (c) and (d) De-Lighter; (e) and (f) ours.

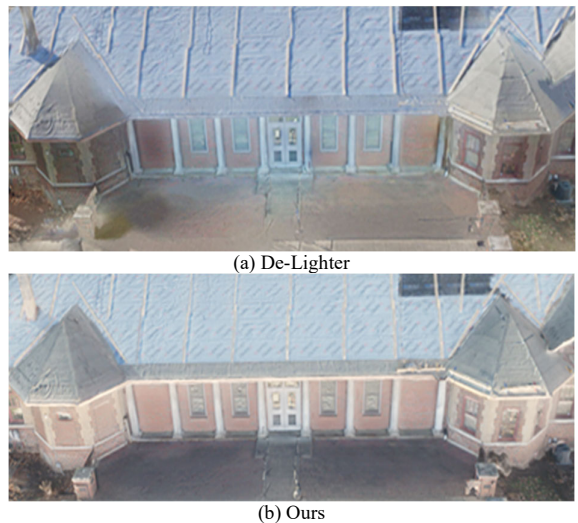

**Figure 14.** Color consistency comparison.

### 4.3 Feature Matching

For the following two experiments, we discussed the contributions of our albedo images to a typical photogrammetric data processing. Since our albedo images are illumination-free, we expect that less illumination variation should result in better feature matching with the current pipeline. Figure 15 shows the feature matching test with epipolar constraint to remove the anomaly, with the SIFT feature extraction and matching functions of OpenCV. The results show that with our albedo images, the extracted and matched features are much better distributed over the image.

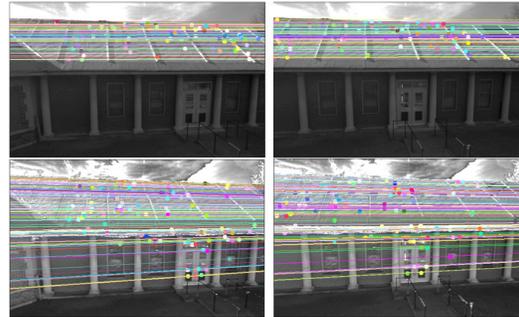

**Figure 15.** Feature matching with epipolar constraint. (Top: with original images, bottom: our albedo images)

We then evaluated the performance on the entire image dataset by using feature matching and bundle adjustment functions of Agisoft Metashape Professional and list metrics of optimization in Table 1. It shows that in all metrics, the albedo images incur better accuracy.

| Metric | Original | Ours |
|---|---|---|
| Matching Inlier Rate | 77.45 % | **80.98 %** |
| RMS Reprojection Error | 1.45 pixels | **1.26 pixels** |
| Max Reprojection Error | 125.94 pixels | **81.13 pixels** |

**Table 1.** Metric of bundle adjustment.

### 4.4 Dense Image Matching

In this experiment, we compared the dense matching quality of original images and our images. Figure 16(c) shows point clouds of pairs with original images Figure 16(a) and albedo images Figure 16(b). It is obvious that our albedo image pair yields more details in the shadow region than the original. Figure 17 presents another experiment in which we manually increase the brightness of original images, to verify that the gain of details is not attributed to brightness changing.

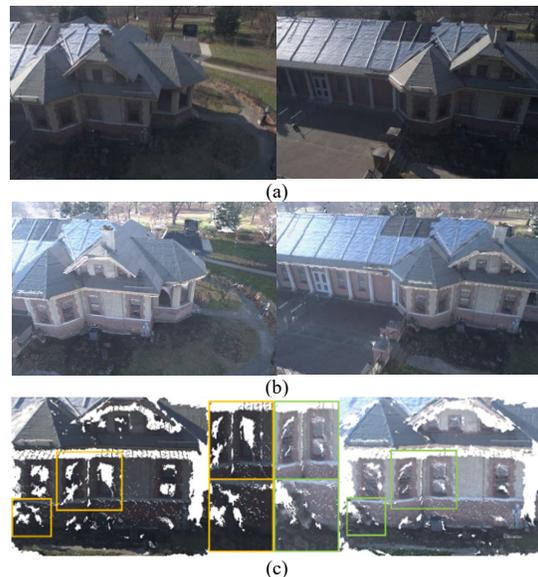

**Figure 16.** Comparison of dense matching point clouds of origin images (a) with our albedo images (b). Left of (c) is generated from original images, right is generated from ours.

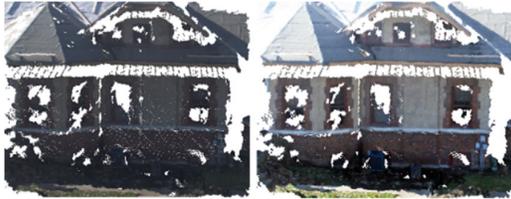

**Figure 17.** Point clouds of original (left) and manually edited images (right).

## 5. CONCLUSION

We present a novel outdoor aerial image formation model and derived albedo recovering method based on the proposed model, and demonstrate with various photogrammetry applications including relighting, feature matching, and dense matching.

Similar to most multi-view intrinsic image decomposition, our approach requires RAW images with linear color space to ensure the correctness of radiometric equations. That means our method cannot directly apply to existing datasets taken in JPEG format. Compared with other multi-view intrinsic image decomposition methods, the proposed method overcomes the difficulty of the lack of direct observation of the environmental radiance, instead, our approach can estimate the lighting condition from images. Thus the proposed method does not require additional user assistance or extra data collection, and our method works on single temporal datasets.

Regarding albedo recovering, compared with the state-of-the-art data-driven methods, our approach demonstrates outstanding performance on correctness and multi-view consistency. The relighting capacity also exceeds the commercial software. We also investigate the possibility of improving geometric quality with our albedo images. It turns out albedo images could refine feature matching and dense matching to improve the geometry.

## ACKNOWLEDGEMENT

This work is supported by the Office of Naval Research (Award No. N000141712928).This work is supported by the Office of Naval Research (Award No. N000141712928).